\pdfoutput=1

\documentclass[11pt]{article}

\usepackage[final]{acl}

\usepackage{times}
\usepackage{latexsym}
\usepackage{algpseudocode}
\usepackage{algorithm}
\usepackage{amsfonts}
\usepackage{multirow}
\usepackage[acceptedWithA]{}

\usepackage[T1]{fontenc}

\usepackage[utf8]{inputenc}

\usepackage{graphicx}

\usepackage{microtype}

\usepackage{inconsolata}

\usepackage{microtype}
\usepackage{multirow}
\usepackage{latexsym}
\usepackage{booktabs}
\usepackage{courier}
\usepackage{amsmath}
\usepackage{amsfonts} 
\usepackage{subfig}
\usepackage{placeins}
\usepackage{tikz}
\usepackage{hyperref}

\usepackage{soul}
\usepackage{color}

\title{A Symbolic Framework for Evaluating Mathematical Reasoning \\and Generalisation with Transformers}

\author{Jordan Meadows$^1$, Marco Valentino$^2$, Damien Teney$^2$, Andr\'e Freitas$^{1,2}$ \\ $^1$University of Manchester, United Kingdom \\  $^2$Idiap Research Institute, Switzerland \\
\texttt{jordan.meadows@postgrad.manchester.ac.uk} \\
\texttt{\{marco.valentino, damien.teney, andre.freitas\}@idiap.ch}}

\begin{document}
\maketitle
\begin{abstract}
This paper proposes a methodology for generating and perturbing detailed derivations of equations at scale, aided by a symbolic engine, to evaluate the generalisability of Transformers to out-of-distribution mathematical reasoning problems. Instantiating the framework in the context of sequence classification tasks, we compare the capabilities of GPT-4, GPT-3.5, and a canon of fine-tuned BERT models, exploring the relationship between specific operators and generalisation failure via the perturbation of reasoning aspects such as symmetry and variable surface forms. Surprisingly, our empirical evaluation reveals that the average in-distribution performance of fine-tuned models surpasses GPT-3.5, and rivals GPT-4. However, perturbations to input reasoning can reduce their performance by up to 80 F1 points. Overall, the results suggest that the in-distribution performance of smaller open-source models may potentially rival GPT by incorporating appropriately structured derivation dependencies during training, and highlight a shared weakness between BERT and GPT involving a relative inability to decode indirect references to mathematical entities. We release the full codebase, constructed datasets, and fine-tuned models to encourage future progress in the field\footnote{\url{https://github.com/jmeadows17/transformers-for-calculus}}.
\end{abstract}

\noindent Out-of-distribution generalisation in Transformers~\citep{vaswani2017attention} is a fundamental and desirable property~\citep{schlegel2023survey,belinkov2022probing,teney2020value}, especially in domains that require rigorous and controlled reasoning such as mathematics, physics, biomedicine, and software verification~\citep{frieder2023mathematical,lee2022establishment,valentino2022hybrid, lewkowycz2022solving,Drori_2022,welleck2021towards,kumar2020adversarial}. Various strategies have been proposed to evaluate model generalisability, including direct input manipulation~\citep{rozanova2023estimating,stolfo2022causal,nie2020adversarial,kaushik2019learning,welleck2022symbolic} and probing on the internal representation ~\citep{rozanova2023interventional,ravichander2021probing,elazar2021amnesic,veitch2020adapting}. However, the adoption of such methods for evaluating generalisation on complex, multi-step reasoning problems is still limited.
Current interventional approaches are challenged by the difficulty of isolating confounding factors, and formalising the expected causal mechanisms that underpin models' predictions~\citep{rozanova2023estimating,stolfo2022causal,ribeiro-etal-2020-beyond,kaushik2019learning}. Particularly in the mathematical domain, these hurdles impact the scope and reliability of causality and robustness studies~\citep{pearl2009causal,shreya2022survey}.

\begin{figure}
    \centering    \includegraphics[width=1.05\columnwidth]{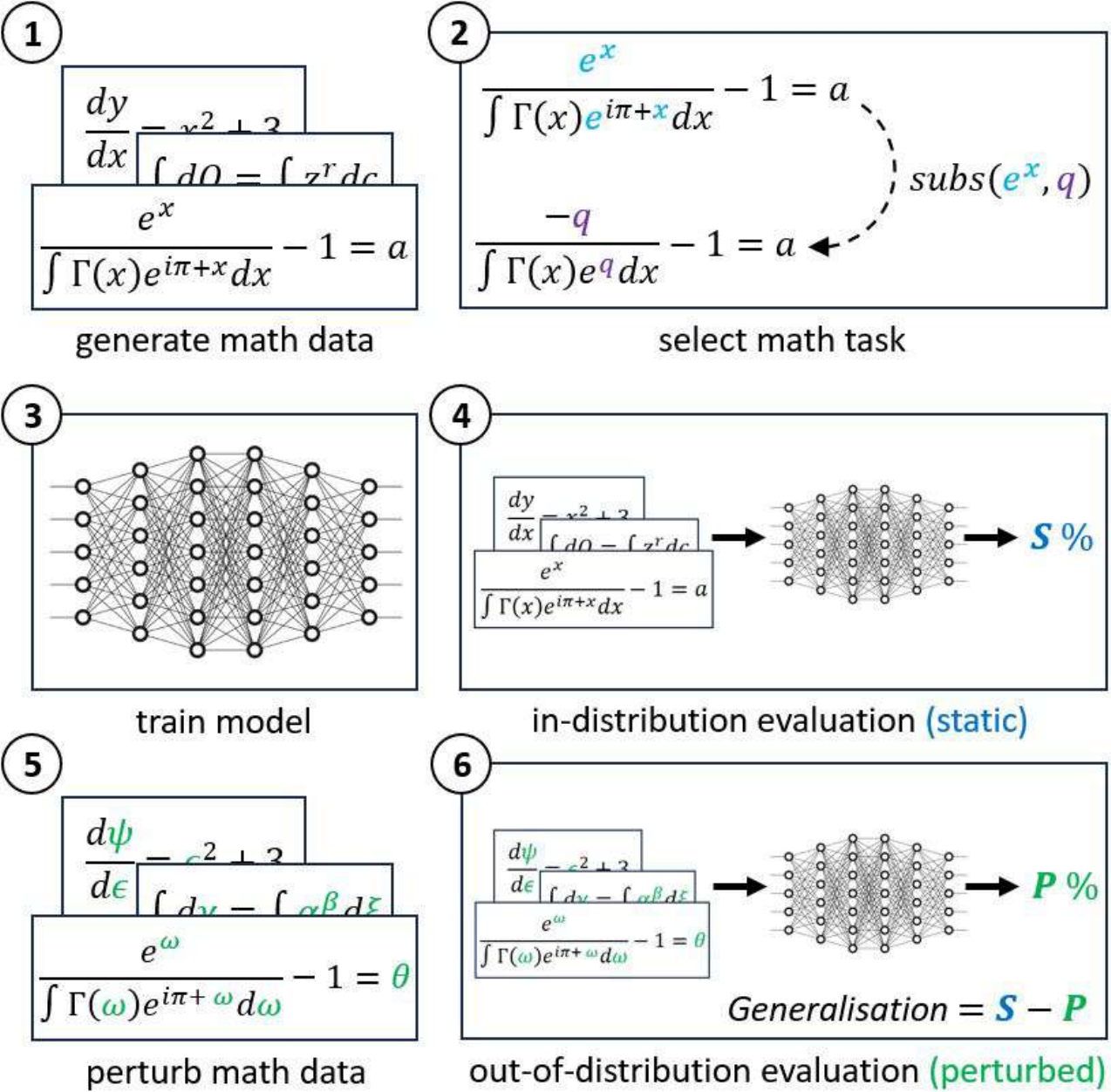}
    \caption{We present a framework for generating and perturbing high-quality mathematical derivations at scale to systematically evaluate mathematical reasoning and generalisation in Transformers.}
    \label{fig:framework}
\end{figure}

We leverage the rich environment of symbolic engines to design a data generation and evaluation framework that can produce high-quality mathematical reasoning steps possessing diverse symbolic properties at scale. In particular, strict symbolic rules offer a systematic approach to \textit{perturbing} mathematical reasoning, and hence evaluating the generalisation of neural models to out-of-distribution textual aspects such as symmetry and variable surface forms. This allows the exploration of deep relationships between semantic and syntactic elements of math reasoning and model generalisability across diverse subdomains, extending beyond the limited interventional scope of previous works~\citep{stolfo2022causal,welleck2022symbolic,patel2021nlp, ribeiro-etal-2020-beyond, kaushik2019learning,yao2021survey}. Additionally, we dialogue with an impending data scarcity problem, where high-quality data is forecast to be outpaced by the training needs of models within the decade~\citep{villalobos2022will}. Symbolic engines~\cite{meurer2017sympy,wolfram1999mathematica} facilitate the generation of annotated mathematical reasoning, which allows the construction of high-quality datasets for various tasks. We combine 18 symbolic operators with rules that guide the exploration of equational state spaces and generate derivations, then perturb and adapt them for exemplar entailment tasks. In this case, these are sequence classification tasks that focus on operator usage in reasoning chains.

To demonstrate our approach, we fine-tune a canon of BERT-based models used in mathematical language processing~\citep{li2023bert, mcnichols2023algebra,meadows2023introduction,zhong2022evaluating}, and test in-context learning methods with GPT-3.5 and GPT-4, to determine their capacity for recognising coherent math reasoning (within this scope), and to abstract fundamental properties impacting their ability to generalise. To summarise, the paper offers the following contributions:

\noindent \textbf{(1.)} A general approach to generating annotated derivations of controllable complexity levels, involving premise equation generation (Alg.~\ref{alg:premise_generation}) and the sequential application of operators to prior equations to derive new results (Alg.~\ref{alg:derivation_generation}). \newline \textbf{(2.)} A systematic and scalable methodology to perturb various aspects of mathematical data including syntax and semantics, implementing several perturbations for evaluation. \newline \textbf{(3.)} An experimental framework for evaluating the out-of-distribution generalisation of models on mathematical reasoning tasks (Fig.~\ref{fig:framework}). \newline \textbf{(4.)} Example instantiation of the framework involving sequence classification tasks. The generated datasets include static and perturbed derivations totalling over 200K examples. \newline \textbf{(5.)} An extensive evaluation of various BERT-based and GPT models culminating in a discussion relating the limited generalisability of models with respect to key operators and mathematical content.

In short, the proposed mathematical data generation and perturbation approach may be integrated into evaluation frameworks for the purpose of testing model generalisability to specific distribution shifts, such as specific surface forms of equations or operator usage. We apply the framework to demonstrate the brittleness of fine-tuned encoder models, and reveal underlying weaknesses shared by both BERT and GPT.

\section{Related Work}

\textbf{Computer algebra.} SymPy~\citep{meurer2017sympy} is a computer algebra system incorporated within numerous approaches. For example, \citet{chen2022program} solve numerical reasoning tasks including simple math elements such as numbers, by prompting language models to generate SymPy solvable code. \citet{mandlecha2022hybrid} use SymPy to generate data for answering questions ranging from arithmetic to calculus without exploring generalisability aspects. \citet{hu2022enhancing} solve a similar array of problems from a large-scale dataset~\citep{saxton2019analysing}, and test for generalisability to an extrapolation set of problems. \citet{Drori_2022} fine-tune the decoder model, Codex~\citep{chen2021evaluating}, on a dataset of questions from MIT's university-level mathematics courses, generating SymPy solution code. \citet{lample2019deep} train a model to integrate and solve differential equations, but do not explore generalisation~\citep{davis2019use}. \citet{welleck2022symbolic} conduct similar experiments using a single model and a single operator (integration) on a single task. We consider 18 operations, 7 models, multiple tasks, and emphasize multi-step equational reasoning.

\textbf{Reasoning with mathematical language.} Transformers~\citep{saxton2019analysing,clark2020Transformers,rabe2020mathematical} defined the state-of-the-art in multiple subdomains and tasks in mathematical language processing~\citep{azerbayev2023llemma,meadows2023introduction,lewkowycz2022solving, Drori_2022}. Transformer encoder models obtain leading performance in related tasks~\citep{ferreira-etal-2022-integer, lai2022semeval,zhong2022evaluating,peng2021mathbert,valentino-etal-2022-textgraphs, tran2022ijs,reusch2022Transformer, novotny2022combining}. The evaluation of the mathematical capabilities of GPT models, and the comparison between GPT and smaller fine-tuned models when deriving equations, has been considered elsewhere~\citep{meadows2023generating,frieder2023mathematical,azerbayev2023llemma}.

\textbf{Data augmentation, synthetic benchmarks, and evaluation frameworks.} Numerous approaches exist related to evaluating the mathematical and symbolic capabilities and robustness of models~\cite{li2020isarstep}. \citet{stolfo2022causal} perturb elements of math word problems~\citep{liang2022mwp} such as numerical operands of implicit arithmetic operations, and natural language, inspired by related work in causal analysis~\citep{pearl2022direct,christiansen2021causal,patel2021nlp,ribeiro-etal-2020-beyond}. Mirroring other notable work~\cite{welleck2022symbolic}, their approach focuses on one or two task-dependent perturbations. Our approach to generating and perturbing data is largely task-independent, and allows for the complex augmentation of operators, variables, expressions, and equations in multi-hop reasoning chains. INT~\cite{wu2020int} is a similar generation metric more closely aligned with formal theorem proving~\cite{polu2020generative,moura2015lean}. Our work is aligned with computer algebra systems~\cite{meurer2017sympy} and more applied mathematical domains (\textit{e.g.,} physics, engineering), and includes calculus.

\section{Generating and Perturbing Derivations with Symbolic Engines}
\label{sec:general_framework}

The data generation approach outputs and perturbs multi-step reasoning involving step annotations and equations. A model learns mathematical reasoning in the context of a given task, is evaluated on an in-distribution test set, then each element of that set is symbolically perturbed and the difference in model inference due to the perturbation contributes to a generalisability measure for the model (Fig.~\ref{fig:framework}). 

To outline this process, given a vocabulary of symbols $\mathcal{V}$ and set of computer algebra operations $\mathcal{R}$, each set is sampled from to ultimately generate an ordered list of steps $s_i \in \mathcal{D}$, where $\mathcal{D}$ represents the output derivation. An initial reasoning step $s_1 = (\text{premise}, \text{annotation})$ is generated such that $\mathcal{D} = [s_1]$. An operation $r \in \mathcal{R}$ is sampled, which in its most general form accepts two operands (arity 2). The first operand is an equation $s_{j,1}$ from tuple $s_j \in \mathcal{D}$. A suitable secondary variable ($\in \mathcal{V}$), expression, or equation operand $m$ is extracted from $\mathcal{D}$, and the next equation is generated by applying operation $r$ through $s_{i+1, 1} = r(s_{j,1}, m)$. The annotation $s_{i+1,2}$ is also a list containing (most generally) the name of the operation, the equation index, and secondary operand, such that $s_{i+1,2} =[r, j, m']$ (where $m'$ is a variable/expression string or equation index representing operand $m$). Therefore, step $s_{i+1} = (r(s_{j,1}, m), [r, j, m'])$. If $\mathcal{D} = [s_1]$, then $i = j = 1$, and the derivation updates such that $\mathcal{D} = [s_1, s_2]$. This process repeats until the derivation reaches a target length.

Given a specific task (\textit{e.g.,} sequence classification), a derivation $\mathcal{D}$ is adapted to form model input $\mathcal{D}'$, such that a static test set is sampled from the same distribution as the training set or in-context examples. This static set $X$ contains task examples $\mathcal{D}'_i$ and labels. For all $\mathcal{D}'_i \in X$, a perturbation $\mathcal{P}$ is applied to corresponding initial derivation $\mathcal{D}_i$ to form a perturbed derivation $\mathcal{P}(\mathcal{D}_i)$, which is similarly adapted to form an out-of-distribution perturbed task example $\mathcal{P}(\mathcal{D}_i)' \in P$, where $P$ is the out-of-distribution test set corresponding to static set $X$, such that $\mathcal{P}:X \rightarrow P$. 

Now that we have static (in-distribution) and perturbed (out-of-distribution) reasoning pairs $(X_i, P_i)$, for a given model $\mathcal{M}$ we can evaluate its generalisability by comparing its static and perturbed predictions, respectively $\mathcal{M}(X_i)$ and $\mathcal{M}(P_i)$. Together with the respective labels, these outputs allow comparisons between aggregate scores (Tab.~\ref{tab:steps} and \ref{tab:direct}), but also a pair-wise analysis involving more sophisticated logic involving $\mathcal{M}(X_i)$ and $\mathcal{M}(P_i)$, which may include perturbations other than $\mathcal{P}$ (Tab.~\ref{table:pairwise_analysis}). 

This provides a highly controllable symbolic framework for evaluating mathematical generalisation capabilities of models. The approach produces mathematical reasoning at scale, and can both improve the depth of mathematical corpora through the generation of underrepresented reasoning, and serve as the backbone for testing model generalisability in numerous settings.

\subsection{Premise Generation}


\noindent To generate premise equations we use a vocabulary (uppercase and lowercase English characters, excluding \{i, e, d, O\}) and a set of 18 operators defined within the symbolic engine, separated by arity. For instance, arity ``0'' represents the introduction of a premise. Arity 1 includes operations that only accept a single variable, expression, or equation (\textit{e.g.,} simplification). Arity 2 includes those such as addition and integration. 

Alg.~\ref{alg:premise_generation} (Appendix~\ref{sec:premise_generation}) describes how operators are recursively applied to vocabulary elements and expressions to produce premises such as: $z(n,f) = f + n$, \; $J(p,w) = e^{p^w}$, \; and \; $Q(x) = \log(x)$. To give a brief description, a symbol is first sampled from the vocabulary. Then an operator with a specific arity is selected, which is applied to the symbol (potentially after selecting another symbol depending on the arity) to form the RHS of an equation. If $\mathcal{C} = 1$, this current RHS will feature as the final RHS of the premise. Otherwise, operators will be re-selected and re-applied to the current RHS up to $\mathcal{C}-1$ times. Once the premise has reached a sufficient complexity, a unique symbol is sampled for the equation LHS, which represents a function of the RHS variables, and the LHS and RHS are conjoined as a SymPy equation.

\subsection{Derivation Generation}

\begin{algorithm}[ht!]
\caption{Generate Derivation Step}
\begin{algorithmic}[1]

\Procedure{Step}{$\mathcal{D}$}
\State Initialise operator sets $\mathcal{R}_0$, $\mathcal{R}_1$, and $\mathcal{R}_2$, and set sampling probabilities
\State Sample arity $a \in \{0, 1, 2\}$
\If {$a = 0$}
\State Sample operator $R \in \mathcal{R}_0$
\State Generate equation using $R$, annotation
\ElsIf {$a = 1$}
\State Sample unary operator $R \in \mathcal{R}_1$
\State Sample equation $e$ from $\mathcal{D}$
\State Generate equation $R(e)$ and annotation
\ElsIf {$a = 2$}
\State Sample binary operator $R \in \mathcal{R}_2$
\State Sample equation $e$ from $\mathcal{D}$
\State Sample variable, expression, or equation, $m$, from $\mathcal{D}$ 
\State Generate equation $R(e,m)$ and annotation
\EndIf
\State \textbf{return} (equation, annotation) \textbf{if} equation is valid \textbf{else} None
\EndProcedure
\end{algorithmic}
\label{alg:step}
\end{algorithm}

\noindent To generate a derivation, a premise equation generated by Alg.~\ref{alg:premise_generation} and an annotation are initially stored as a tuple (equation, annotation) within a list $\mathcal{D}$. This list is input to the \texttt{Step} procedure (Alg.~\ref{alg:step}) which considers any mathematical elements defined in $\mathcal{D}$ so far, and uses them as input to the operators to generate new (equation, annotation) tuples. Tuples containing equations that pass validity checks such as maximum character length (250), LHS/RHS redundancies (\textit{e.g.,} $x=x$), and checks related to integration (etc.), are appended to $\mathcal{D}$. A final derivation is outputted when $\mathcal{D}$ exceeds a target length. This process is described further in Alg.~\ref{alg:derivation_generation}, including a description of its output rate ($7$ seconds/step on mid-range CPU), hyperparameters, and equation sampling, in Appendix~\ref{sec:derivation_generation}.

\subsection{Perturbations}

\begin{figure*}[h!]
    \centering
    \includegraphics[width=1\textwidth]{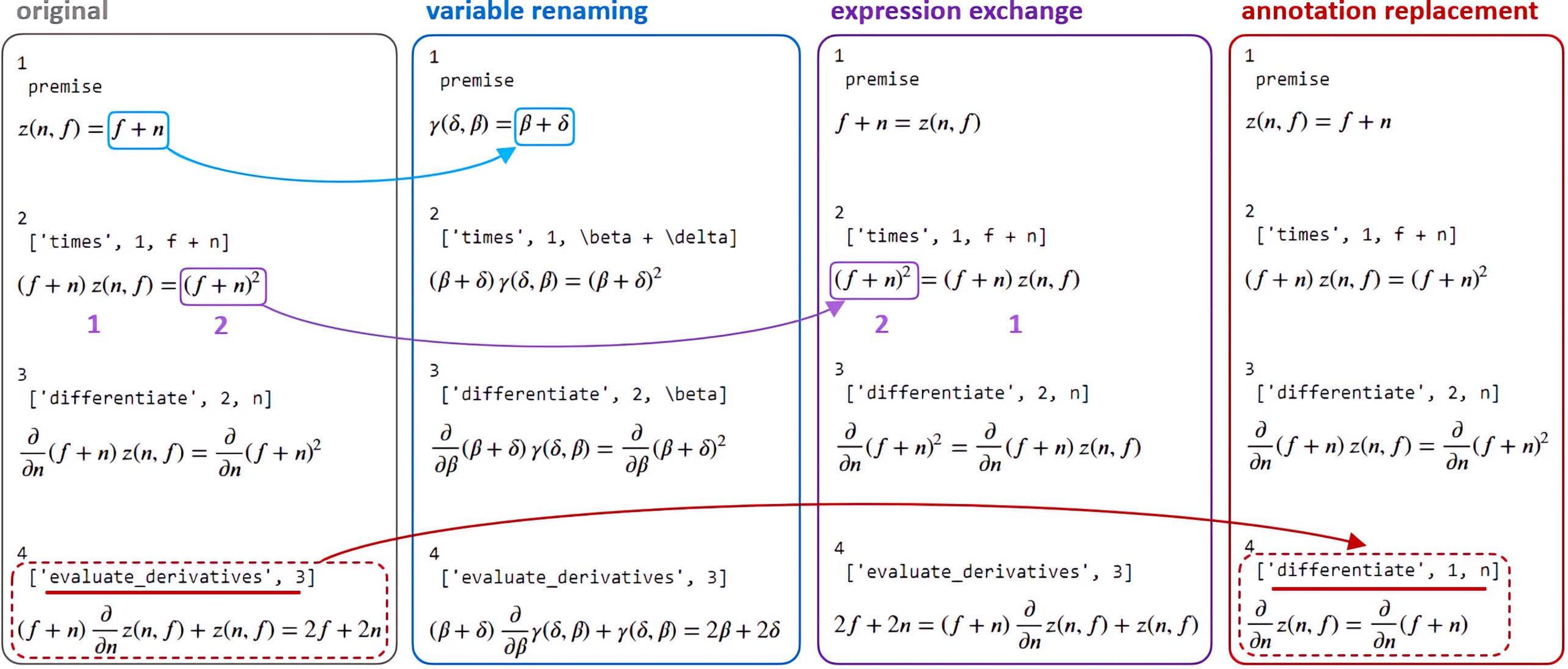}
    \caption{Example perturbations applied to a generated derivation using computer algebra.}
    \label{fig:perturbations}
\end{figure*}

To perturb LaTeX sequences, the examples in the static set are \textit{re-interpreted} by the computer algebra system using SymPy's \texttt{srepr} tree representation. In this paper, we consider four different perturbations for evaluation (Fig.~\ref{fig:perturbations}). However, the compatibility with the computer algebra system allows a wide variety of perturbations that range from small-scale interventions to single variables, through to long-range interventions that target complex semantic relationships between any number of distant sequence elements. For instance, one may choose to only perturb reasoning chains that involve a premise renaming operation followed directly by integration, or square a variable and propagate that change through the entire reasoning chain. The perturbations adopted in our evaluation are as follows:

\textbf{Variable Renaming (VR)}. For each example in the static set, we uniquely map each symbol to an out-of-vocabulary symbol sampled from 10 Greek letters (\textit{e.g.,} $E(n, x) = n + x$ becomes $\alpha(\beta, \gamma) = \beta + \gamma$).

\textbf{Expression Exchange (EE).} For each example in the static set, we swap expressions either side of the equality (\textit{e.g.,} $E(n, x) = n + x$ becomes $n + x = E(n, x)$). This reverses the overrepresentation of LHS functions in the static set.

\textbf{Annotation Replacement (AR).} Each example in the static set contains a correct and incorrect final equation. For each example, the operator and operands (and hence the annotation) responsible for generating the negative equation are calculated, replacing the corresponding annotation in the sequence and swapping the label (i.e. from positive to negative and vice-versa).

\textbf{Equation Conversion (EC).} If a sequence consists of a chain such as $\log(x) \text{ [SEP] } x \text{ [SEP] } \frac{1}{x}$, and the implicit operation is differentiation, a random symbol is sampled from the vocabulary (\textit{e.g.,} $Q$), and the sequence becomes $Q(x) = \log(x) \text{ [SEP] } x \text{ [SEP] } \frac{dQ(x)}{dx} = \frac{1}{x}$. If integrating, then the (negative) sequence becomes $\frac{dQ(x)}{dx} = \log(x) \text{ [SEP] } x \text{ [SEP] } Q(x) = \frac{1}{x}$.
 
\section{Sequence Classification Tasks}

The data generation approach (Alg.~\ref{alg:step}-\ref{alg:derivation_generation}) outputs math reasoning which may then be adapted for specific tasks. As such, we instantiate the general framework described in Fig.~\ref{fig:framework} in the context of \textit{two sequence classification tasks}. Task examples, dataset sizes, and model input (Fig.~\ref{fig:input}) are described in Appendix~\ref{sec:data}.

\textbf{Derivation Step Classification.} Given multiple steps of a derivation, such as those in Fig.~\ref{fig:perturbations} and \ref{fig:input}, the final equation has a $50\%$ chance of being replaced with a similar equation that is \textit{associated with a different annotation}, and the model must classify whether the overall derivation is coherent (\textit{i.e.,} whether the annotations match the equations and overall reasoning). Incoherent equations associated with a negative label are generated by applying a different operator, or the same operator with different inputs, to a previous derivation equation through \texttt{Step} (Alg.~\ref{alg:step}). To solve this task a model must learn the necessary \textit{inter-statement dependencies} required to form the final equation in the derivation, guided by the final annotation. These dependencies are a crucial aspect of derivations and equational reasoning.

\textbf{Calculus Classification.} Given a simplified sequence containing only a \textit{premise expression}, a \textit{variable}, and a \textit{final expression} (Fig.~\ref{fig:input}), where the final expression is the premise differentiated or integrated with respect to the variable \newline (\textit{e.g.,} \;\; premise:\; $\log(x)$, \;\; var:\; $x$, \;\; final:\; $1/x$), \newline the final expression has a $50\%$ chance of being swapped with a similar but incorrect expression. The negative examples are generated by differentiating/integrating the dataset premises by fixing the variable and changing the premise, or vice versa. To select the expression for the final expression swap to form the incoherent sequence (negative label), these expressions are ranked by their Damerau-Levenshtein distance~\cite{zhao2019string,meadows2021similarity} from the ground truth. For example, the expression $-T + sin(U)$ is differentiated with respect to $T$ to give $-1$. The expression corresponding to the negative label is $1$.

\section{Evaluation}

{\renewcommand{\arraystretch}{1}%
\begin{table*}
	\centering
        \small
	\scalebox{0.85}{
		\begin{tabular}{@{}lcc|cc|cc|cc|cc|ccc @{}}
            \toprule
			& \multicolumn{2}{c|}{\textbf{Static}} & \multicolumn{2}{c|}{\textbf{VR}} & \multicolumn{2}{c|}{\textbf{EE}} & \multicolumn{2}{c|}{\textbf{AR}} & \multicolumn{2}{c|}{\textbf{s - 1}} & \multicolumn{2}{c}{\textbf{s - 2}}\\
		\midrule
			& Acc & F1 & Acc & F1 & Acc & F1 & Acc & F1 & Acc & F1 & Acc & F1\\
		\midrule
			BERT-base-uncased (\textbf{s=2})   & 87.7 & 88.9 & 87.0 & 88.1 & 87.0 & 88.0 & 87.5 & 88.7 & -  & - & - & - \\
                BERT-base-uncased (\textbf{s=3}) & 78.9 & 78.7 & 71.9 & 71.0 & 69.1 & 66.0 & 53.7 & 50.6 & 68.4 & 69.0 & - & -\\
                 BERT-base-uncased (\textbf{s=4}) & 58.8 & 63.6 & 55.0 & 60.3 & 56.4 & 60.3 & 42.4 & 48.1 & 65.7 & 62.2 & 52.8 & 29.8\\
		\midrule
                 BERT-base-cased (\textbf{s=2}) & 87.2 & 88.5 & 81.9 & 83.2 & 85.3 & 86.1 & 85.5 & 87.2 & - & - & - & -\\
                 BERT-base-cased (\textbf{s=3}) & 78.2 & 77.3 & 68.8 & 64.5 & 65.0 & 58.9 & 54.5 & 49.6 & 54.6 & 30.5 & - & -\\
                 BERT-base-cased (\textbf{s=4}) & 66.8 & 71.7 & 58.5 & 61.5 & 62.6 & \textbf{\textit{\underline{67.2}}} & 43.3 & 53.1 & 71.9 & 73.9 & 54.3 & 21.8\\
		\midrule
                 MathBERT (\textbf{s=2}) & 83.2 & 82.0 & 76.2 & 70.6 & 79.0 & 75.7 & 78.5 & 76.0 & - & - & - & -\\
                 MathBERT (\textbf{s=3}) & 84.2 & 83.9 & 69.1 & 64.5 & 63.3 & 52.2 & 66.3 & 64.0 & 67.4 & 58.7 & - & -\\
                 MathBERT (\textbf{s=4}) & 67.1 & 68.4 & 59.5 & 52.6 & 62.3 & 62.1 & 48.5 & 47.9 & 68.6 & 68.0 & 51.8 & 29.0\\
		\midrule
                  SciBERT-uncased (\textbf{s=2}) & 92.5 & 92.6 & 72.9 & 70.4 & 86.8 & 86.1 & 90.0 & 90.2 & - & - & - & -\\
                   SciBERT-uncased (\textbf{s=3}) & 88.9 & \textbf{\textit{89.4}} & 82.1 & \textbf{\textit{81.9}} & 70.3 & 66.4 & 70.9 & \textbf{\textit{72.2}} & 80.6 & \textbf{\textit{81.8}} & - & - \\
                   SciBERT-uncased (\textbf{s=4}) & 76.3 & \textbf{\textit{\underline{76.5}}} & 69.5 & 66.8 & 68.6 & 65.9 & 60.7 & \textbf{\textit{\underline{59.6}}} & 76.9 & \textbf{\textit{\underline{77.9}}} & 59.3 & 57.4\\
		\midrule
                   SciBERT-cased (\textbf{s=2}) & 92.6 & \textbf{93.1} & 85.3 & 87.1 & 89.8 & \textbf{90.2} & 91.0 & \textbf{91.7} & - & - & - & -\\
                   SciBERT-cased (\textbf{s=3}) & 77.2 & 72.4 & 72.7 & 67.2 & 61.0 & 44.1 & 50.8 & 29.5 & 52.9 & 12.8 & - & - \\
                   SciBERT-cased (\textbf{s=4}) & 71.0 & 70.9 & 65.1 & 64.6 & 66.6 & 65.4 & 47.0 & 42.9 & 77.9 & 74.9 & 52.7 & 11.0\\
		\midrule
                   Encoder Average (\textbf{s=2}) & 88.6 & 89.0 & 80.7 & 79.9 & 85.6 & 85.3 & 86.5 & 86.8 & - & - & - & -\\
                   Encoder Average (\textbf{s=3}) & 81.5 & 80.3 & 72.9 & 69.8 & 65.7 & 57.5 & 59.2 & 53.2 & - & - & - & -\\
                   Encoder Average (\textbf{s=4}) & 68.0 & 70.2 & 61.5 & 61.2 & 63.3 & 64.2 & 48.4 & 50.3 & - & - & - & -\\
            \midrule
                   GPT-3.5 (\textbf{s=2}) & 66.0 & 72.6 & 65.5 &
                   72.5 & 59.0 & 65.3 & 53.0 & 63.3 & - & - & - & -\\
                   GPT-3.5 (\textbf{s=3}) & 
                   57.0 & 64.2 & 61.5 &
                   67.0 & 60.5 & 65.5 &
                   46.0 & 54.2 & 56.5 & 64.5 & - & -\\
                   GPT-3.5 (\textbf{s=4}) &
                   51.5 & 59.1 & 49.5 & 56.3 & 54.0 & 59.6 & 44.5 & 52.8 & 56.0 & 62.7 & 59.0 & 67.7\\
            \midrule
                   GPT-4 (\textbf{s=2}) & 88.0 & 88.5 & 87.5 & \textbf{88.2} & 82.5 & 81.1 & 64.5 & 66.4 & - & - & - & -\\
                   GPT-4 (\textbf{s=3}) & 77.5 & 77.4 & 77.5 & 76.7 & 78.5 & \textbf{\textit{77.2}} & 50.0 & 55.0 & 73.5 & 77.4 & - & -\\
                   GPT-4 (\textbf{s=4}) & 68.0 & 68.0 & 69.0 & \textbf{\textit{\underline{69.6}}} & 66.0 & 64.6 & 42.0 & 42.6 & 76.0 & 76.9 & 77.5 & \textbf{\textit{\underline{80.2}}}\\
		\midrule
                   Encoder (steps avg) & 79.4 & 79.8 & 71.7 & 70.3 & 71.5 & 69.0 & 64.7 & 63.4 & - & - & - & -\\
            \midrule 
                   GPT-3.5 (steps avg)    & 58.2 & 65.3 & 58.8 & 65.3 & 57.8 & 63.5 & 47.8 & 56.8 & - & - & - & -\\
            \midrule 
                   GPT-4 (steps avg) & 77.8 & 78.0 & 78.0 & 78.2 & 75.7 & 74.3 & 52.2 & 54.7 & - & - & - & -\\
            \midrule
		\end{tabular}
		}
	\caption{\textit{Derivation Step Classification} task results. Bold numbers denote highest F1 scores for \textbf{2-step} derivations. Bold italic numbers denote highest \textbf{\textit{3-step}} scores. Bold, italic, and underlined numbers denote highest \textbf{\textit{\underline{4-step}}} scores.}
	\label{tab:steps}
\end{table*}}

In this section, we discuss how the scores obtained by the BERT and GPT models reflect their reasoning capabilities within the scope of the classification tasks described in the previous section.

\textbf{Training and prompts.} Details related to fine-tuning BERT and prompting GPT are given in Appendix~\ref{sec:training}. To summarise, for a single pre-trained BERT model, \textit{five} fine-tuned models are trained for each of the task variations. For instance, in \textit{Derivation Step Classification}, a model is trained \textit{per number of steps} in the input derivations (\textit{i.e.,} 2, 3, and 4 steps). For \textit{Calculus Classification} a model is trained \textit{per operation} (\textit{i.e.,} differentiation and integration). The GPT models are given 4-shot prompts. While we acknowledge that chain-of-thought~\cite{wei2022chain} and tree-of-thought~\cite{yao2023tree} methods may lead to improved mathematical inference, we instead rely on a simple few-shot prompt (Tab.~\ref{table:prompt_exploration}) and focus on the evaluation framework and data generation pipeline. 

\textbf{Further generalisation.} In addition to the out-of-distribution perturbations applied to input sequences for each task, a model that can sufficiently generalise the underlying reasoning should be able to solve (on average) mathematically less complex versions of problems encountered during training. In \textit{Derivation Step Classification}, we evaluate models trained on derivations with a fixed step count on a set of derivations composed of a lower number of steps. This is represented in the $\textbf{s - 1}$ and $\textbf{s - 2}$ columns in Tab.~\ref{tab:steps} given initial step count, $\textbf{s}$. In \textit{Calculus Classification}, where models are exposed to examples comprising \textit{at least two} variables, (\textit{e.g.,} $\cos(ax) - z$) we generate a set of easier problems with 1.5k examples consisting of only one variable (\textit{e.g.,} $\cos(x)$), corresponding to the \textbf{Easy} column of Tab.~\ref{tab:direct}. These out-of-distribution datasets complement the perturbations in the following discussion.

{\renewcommand{\arraystretch}{1.2}%
\begin{table}[h!]
	\centering
        \small
	\scalebox{0.7}{
		\begin{tabular}{@{}lcc|cc|cc|ccccccc @{}}
                \toprule
			& \multicolumn{2}{c|}{\textbf{Static}} & \multicolumn{2}{c|}{\textbf{VR}} & \multicolumn{2}{c|}{\textbf{EC}} & \multicolumn{2}{c}{\textbf{Easy}}\\
			\midrule
			& Acc & F1 & Acc & F1 & Acc & F1 & Acc & F1\\
			\midrule
			BERT-base-uncased (\textbf{int})   & 90.0 & 90.7 & 68.8 & 70.4 & 75.1 & 78.0 & 62.7 & \textbf{72.9}\\
            BERT-base-uncased (\textbf{diff})   & 75.9 & 80.3 & 64.9 & 73.3 & 62.2 & \textbf{\textit{69.8}} & 55.1 & 69.1\\
			\midrule
            BERT-base-cased (\textbf{int}) & 93.0 & 93.4 & 71.6 & \textbf{77.7} & 85.2 & \textbf{86.7} & 63.8 & 71.8\\
            BERT-base-cased (\textbf{diff})  & 74.2 & 77.9 & 64.2 & 72.4 & 60.3 & 64.9 & 56.7 & \textbf{\textit{69.6}}\\
			\midrule
            MathBERT (\textbf{int})   & 92.2 & 92.3 & 74.4 & 75.8 & 74.4 & 71.8 & 58.6 & 68.6\\
            MathBERT (\textbf{diff})   & 84.7 & 85.9 & 59.7 & 48.1 & 58.4 & 47.3 & 56.1 & 50.0\\
			\midrule
            SciBERT-uncased (\textbf{int})  & 96.8 & 96.8 & 65.6 & 74.4 & 54.1 & 15.8 & 62.6 & 71.1\\
            SciBERT-uncased (\textbf{diff}) & 91.8 & 92.3 & 72.6 & \textbf{\textit{76.5}} & 66.8 & 58.1 & 55.2 & 67.8\\
			\midrule
            SciBERT-cased (\textbf{int})  & 97.1 & \textbf{97.2} & 68.1 & 75.8 & 54.2 & 17.0 & 58.0 & 67.1\\
            SciBERT-cased (\textbf{diff}) & 92.3 & \textbf{\textit{92.7}} & 70.9 & 76.5 & 65.4 & 54.6 & 61.5 & 72.3\\
			\midrule
             Encoder Average (\textbf{int})  & 93.8 & 93.2 & 69.7 & 74.8 & 68.6 & 53.7 & 61.1 & 70.3\\
            Encoder Average (\textbf{diff}) & 83.8 & 85.8 & 66.5 & 69.4 & 62.6 & 58.9 & 56.9 & 65.8\\
			\midrule
            GPT-3.5 (\textbf{int}) & 49.5 & 56.3 & 49.5 & 56.3 & 51.5 & 60.1 & 54.5 & 58.1\\
            GPT-3.5 (\textbf{diff}) & 49.0 & 55.3 & 48.5 & 54.2 & 53.0 & 65.7 & 54.5 & 59.2\\
            \midrule
            GPT-4 (\textbf{int}) &
            64.0 & 60.0 & 67.0 & 64.1 & 66.5 & 68.5 & 57.5 & 56.4\\ 
            GPT-4 (\textbf{diff}) & 59.5 & 55.2 & 61.0 & 57.1 & 66.5 & 72.9 & 68.5 & 66.3\\
            \midrule
            Encoders (int/diff avg) & 88.8 & 89.5 & 68.1 & 72.1 & 65.6 & 56.3 & 59.0 & 68.1\\
                \bottomrule
		\end{tabular}
		}
	\caption{\textit{Calculus Classification} task results. Bold numbers denote highest F1 scores for \textbf{integration} derivations. Bold italic denotes highest \textbf{\textit{differentiation}} scores.}
	\label{tab:direct}
\end{table}} 

\textbf{GPT-4 rivals in-distribution performance of fine-tuned BERT-based models while demonstrating better generalisation.} Assuming a suitably descriptive few-shot prompt, where necessary context is provided through either the task description or in-context examples (Appendix~\ref{sec:training}), GPT-4 can rival the average static scores of the fine-tuned encoder models, and surpass them on out-of-distribution test sets, \textit{even without chain-of-thought prompting}. This is shown by the \textit{Derivation Step Classification} results (Tab.~\ref{tab:steps}). For instance, SciBERT-cased (\textbf{s=4}) scores 11\% F1 when classifying sequences with \textbf{s=2} steps. GPT-4 obtains 80\% F1 in this case. Similar generalisation is observed on the \textbf{VR} (\textit{Variable Renaming}) perturbation data, likely due to GPT-4's exposure to vast vocabularies of mathematical symbols (\textit{e.g.,} Greek symbols), and the \textbf{EE} (\textit{Expression Exchange}) set, likely due to GPT-4's exposure to equations with RHS functions which lessens the impact of LHS function bias.

\textbf{GPT-4 can fail to predict mathematical coherence from in-context examples alone.} The \textit{Calculus Classification} task includes minimalistic sequences without operation annotations. Surprisingly, while GPT-4 achieves the best performance on \textit{Derivation Step Classification}, competitive performance is not observed in \textit{Calculus Classification} despite its lower complexity. We attribute this to the fact that, unlike BERT, GPT is not fine-tuned on a specific operation, and in-context examples alone might not contain enough information to consistently discriminate whether a particular sequence involves either differentiation or integration. This is evidenced by the fact that both GPT models score \textit{higher} on the \textbf{EC} (\textit{Equation Conversion}) set. The EC perturbation changes nothing about the operation being performed, but \textit{adds context} by writing (\textit{e.g.}) differentiated expressions as equations with a LHS that includes $\frac{d}{dx}$. F1 scores in GPT models \textit{increase by up to 12 points} in this case, while BERT-based scores \textit{decrease by up to 80 points}  (Tab.~\ref{tab:direct}). To reinforce this, in \textit{Derivation Step Classification}, both GPT models obtain comparatively lower scores on the \textbf{AR} (\textit{Annotation Replacement}) set. This is because sufficient context has been provided only for an operator that differs to the main annotation operator. GPT only learns the format of the sequences and the expected output for the task in this case. However, static performance is maximised by designing the prompt in this manner (Tab.~\ref{table:prompt_exploration}).

\textbf{GPT-3.5 cannot effectively classify mathematical reasoning.} GPT-3.5 scores \textit{15 less} F1 points than the average encoder score of 80\% on the in-distribution set, and is notably outperformed by BERT-based models on most test sets (particularly SciBERT). A notable exception are those that contain less steps (Tab.~\ref{tab:steps}), where performance generally \textit{increases} comparative to static in-distribution scores. This contrasts with the significant corresponding performance drops observed in the BERT-based evaluation, indicating that GPT learns enough from in-context examples to generalise to derivations with less steps, and therefore has a deeper relative understanding of the underlying mathematics.

\textbf{Encoder models fail to generalise.} For \textit{Derivation Step Classification}, models average $80\%$ F1 over all static derivation lengths, and decreases due to perturbations average $10\%$ (\textbf{VR}), $11\%$ (\textbf{EE}), and $16\%$ (\textbf{AR}). This is at most $4\%$ above F1 majority baseline. BERT-uncased and SciBERT-cased fine-tuned on 2-step derivations are exceptions, but the 13 other encoder models are sensitive to at least one perturbation. All models tested do not generalise to \textit{less} derivation steps, reaching as low as $11\%$ F1. In \textit{Calculus Classification} static scores average $90\%$ and perturbations decrease this by $17\%$ (\textbf{VR}) and $33\%$ (\textbf{EC}). All fine-tuned models fail to generalise to perturbations and simpler examples, \textit{with $97\%$ F1 scores repeatedly dropping below $17\%$}. Despite the in-distribution performance, this indicates their reliance on superficial patterns rather than the underlying rules of the operators.

\subsection{Relating Operators to Model Generalisability via Pairwise Analysis}

{\renewcommand{\arraystretch}{1.4}%
\begin{table*}[h!]
\centering
\small
\scalebox{1}{
\begin{tabular}{ c | c | c | c | c } 
  & \textbf{Static} $(S)$ & \textbf{Generalisability} $(G)$ & \textbf{None} & \textbf{All}\\ [0.1ex] 
 \hline
 \multirow{2}{*}{BERT} & 76.0 & 3.3 & 16.5 & 60.8\\
 & $\int_E \; R \; \int \; \partial \; \times$ &
 $\int_E \; R \; + \; \partial_E \; -$ &
 $S_L \; S_R \; + \; X^O \; \times$ &
 $\int \; \partial \; \times \; - \; X^O$\\
 \hline
 \multirow{2}{*}{MathBERT} & 79.7 & 9.0 & 13.2 & 57.2\\
 & $\int_E \; R \; \int \; \partial \; \partial_E$ &
 $R \; \int_E \; X^O \; \partial_E \; \div$ &
 $+ \; S_L \; \div \; S_R \; \cos$ &
 $\partial \; \int \; X^O \; + \; \div$\\
 \hline
 \multirow{2}{*}{SciBERT} & \textbf{87.8} & 5.0 & \textbf{7.0} & 62.7\\
 & $R \; \int_E \; \int \; - \; \div$ &
 $R \; \div \; \partial_E \; + \; X^O$ &
 $S_L \; S_R \; + \; \cos \times$ &
 $\int \; \partial \; - \; + \; \partial_E$\\
 \hline
 \multirow{2}{*}{GPT-3.5} & 58.2 & 2.3 & 29.7 & 45.5\\
 & $\cos \; X^O \; \partial \; \int \; R$ &
 $S_L \; \int_E \; S_R \; + \; X^O$ &
 $- \; \int_E \; \times \; + \div$ &
 $\cos \; X^O \; \int \; \partial \; \partial_E$\\
 \hline
 \multirow{2}{*}{GPT-4} & 77.8 & \textbf{1.7} & 12.0 & \textbf{64.7}\\
 & $\cos \; \partial \; \int \; X^O \; \int_E$ &
 $\cos \; \times \; \partial_E \; \div \; R$ &
 $S_L \; S_R \; - \; R \; \times$ &
 $\cos \; \partial \; X^O \; \int \times$\\
 \hline
\end{tabular}}

\caption{\textbf{Static} $(S)$ represents model accuracy with respect to unperturbed examples. \textbf{Generalisability} $(G)$ represents the percentage of examples where static predictions are correct and all perturbed predictions failed (lower is better). \textbf{None} represents examples where models failed predictions in all cases, and \textbf{All} represents the opposite. Symbols correspond to the top-5 most frequent (final) operators in each unperturbed sequence, where frequency is normalized with respect to operator count in the static set. $R$ is a premise renaming operator. $\int$ and $\partial$ are integration and differentiation operators. $\int_E$ and $\partial_E$ are respective evaluation operators. $X^O$ is exponentiation, $S_L$ and $S_R$ are LHS and RHS substitutions, and arithmetic symbols have their usual meaning.} 
\label{table:pairwise_analysis}
\end{table*}}

\begin{figure*}[h!]
    \centering
    \includegraphics[width=0.9\linewidth]{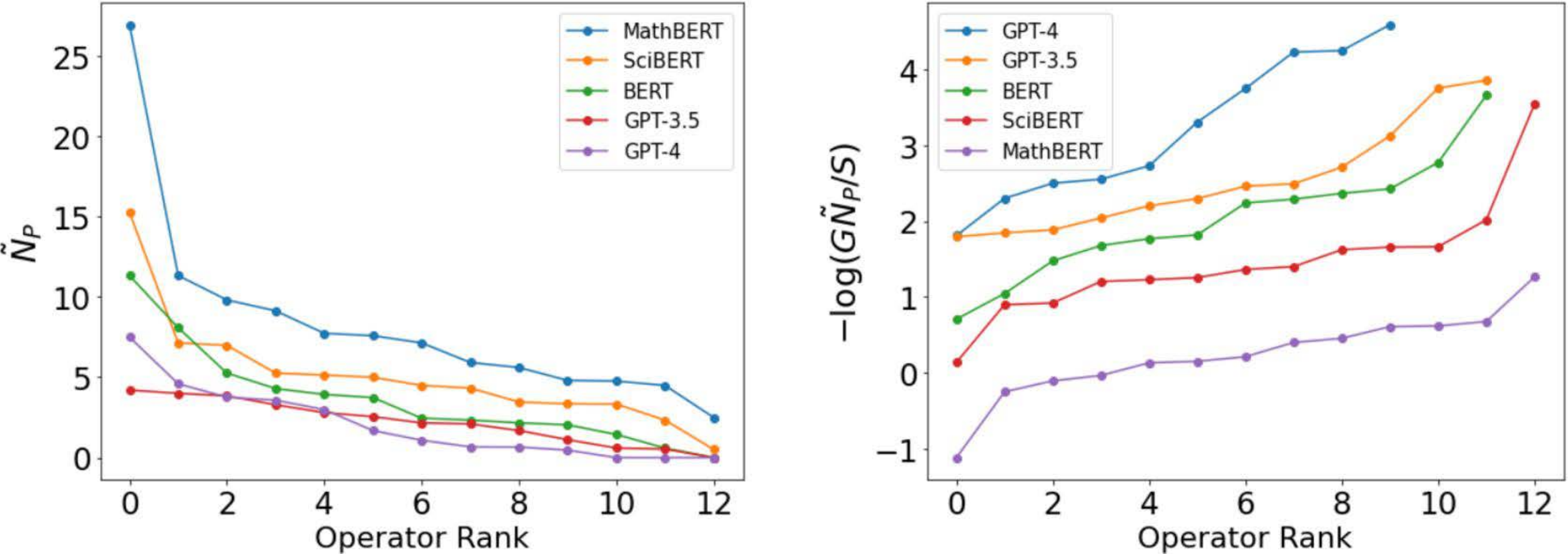}
    \caption{$\tilde{N}_P$ is the percentage of operators present in examples where models fail to generalise to perturbations. The leftmost displays how this proportion varies as a function of operator rank. The rightmost graph factors in static performance $(S)$ and generalisability $(G)$ scores for a clearer comparison of models.}
    \label{fig:operators}
\end{figure*}

In addition to the above evaluation, because the framework offers in-distribution and out-of-distribution pairs that correspond to a single reasoning chain and its perturbation (Fig.~\ref{fig:perturbations}), we can measure generalisability using alternative methods. These are discussed in Tab.~\ref{table:pairwise_analysis}, and further discussed in Appendix~\ref{sec:qual_mat}, but we summarise our findings here.

\textbf{Which operators are most difficult to learn?} Substitution is dependency-wise the most complicated operation and is not associated with a fixed token (such as addition's "+"). It requires a deeper understanding of derivation structure due to a necessary reliance on dependency relations across equations. All models interpret substitution relatively poorly (\textbf{None} column, Tab.~\ref{table:pairwise_analysis}). Operator usage that is easier for models to recognise (and generalise) involves integration or differentiation (\textbf{All} column, Tab.~\ref{table:pairwise_analysis}), and these are associated with specific text spans such as "\textbackslash int" or "\textbackslash partial". Together, this indicates that all models struggle most when operators are \textit{not associated with fixed text spans} or when they rely on \textit{explicitly structured dependency relations}. To give further examples, the fixed text span associated with the addition operator is ``+'', and structured dependency relations are given by the substitution operator's reference to prior equation indexes.

\textbf{Which operations contribute to poor generalisability?} We consider the proportion of examples where static predictions succeed while all perturbation predictions fail (column $G$, Tab.~\ref{table:pairwise_analysis}). For BERT models, \textit{premise renaming} and integration/differentiation \textit{evaluation} operations rank highly, yet this is not mirrored by GPT. Fig.~\ref{fig:operators} explains this difference, displaying the proportion of operators ($\tilde{N}_P$) that contribute to examples where models generalise poorly at a given rank. For example, the highest ranking operator for MathBERT has $\tilde{N}_P > 25$. From Tab.~\ref{table:pairwise_analysis} this operator performs \textit{premise renaming}, denoted by $R$, and over 1/4 of examples involving $R$ contribute to poor model generalisability. In fact for all BERT-based models, the $R$ (and less so the int/diff evaluation) operators have a higher $\tilde{N}_P$ than others. This effect is less prominent for the GPT models. This indicates that high ranking operators have a major impact on generalisation in BERT models, but it is likely that other factors (such as the complexity of equations) are more impactful for GPT. 

\section{Conclusion}

This research presents an approach for generating synthetic data, and a framework for evaluating the mathematical capabilities of models, which may be utilised for purposes beyond sequence classification~\cite{valentino2023multi,meadows2023generating}. We discover that inference failures occur for BERT and GPT models when tasks require complex indirect textual references and inter-statement dependencies. This demonstrates how transformer-based models fail to appropriately infer structured information from linear text. Our findings reveal the complete generalisation failure of BERT models to simple perturbations despite their continued use in the mathematical domain, yet, the experiments reveal that BERT-related models may outperform or match few-shot GPT performance in math classification tasks despite the disparity in pre-training efforts and parameter count. We also observe that perturbations (\textit{e.g.,} EC) which increase the depth of mathematical operator trees while introducing useful task context improves few-shot performance, yet transformers clearly struggle if the underlying dependency graphs of mathematical sequences are too complex. Overall, this paper underscores the potential of using symbolic engines to generate extensive, high-quality mathematical datasets that may be used to explore model weaknesses, and improve mathematical reasoning and generalisation in quantitative domains.

\section{Limitations}

\paragraph{Overall ethical impact.} This work explores a systematic way to elicit the mathematical/symbolic inference properties of Transformer-based models in mathematical language processing tasks. As such, it contributes in the direction of a critique of the reasoning capabilities and the biases of these models.

\paragraph{Chain-of-Thought.} Chain-of-thought and related prompt engineering may lead to improved LLM-based mathematical reasoning. This paper focuses on the exploration and application of the evaluation framework, rather than maximising the mathematical proficiency of language models. 

\paragraph{Derivation generation.} There are irrelevant steps in some derivations, such as applying an operation to an equation but not using the result. This should not affect results as the final equation always depends on a previous equation (except when it is the result of a premise selection operation). This limitation stems from incorrect subderivation extraction from longer chains and will be improved.

\paragraph{Integration.} SymPy does not generate integration constants. Although we account for this within derivation generation, we currently exclude the evaluation of double (or above) integrals, and do not introduce an integration constant when generating expressions for the Calculus Classification task. 

\section*{Acknowledgements}

This work was partially funded by the Swiss National Science Foundation (SNSF) project NeuMath (\href{https://data.snf.ch/grants/grant/204617}{200021\_204617}), the EPSRC grant EP/T026995/1 entitled “EnnCore: End-to-End Conceptual Guarding of Neural Architectures” under Security for all in an AI enabled society, the CRUK National Biomarker Centre, and supported by the Manchester Experimental Cancer Medicine Centre and the NIHR Manchester Biomedical Research Centre.

\bibliography{custom}

\appendix

\section{Fine-tuning BERT and prompting GPT}
\label{sec:training}

\textbf{Fine-tuning BERT.} Transformer encoders with a binary sequence classification layer are fine-tuned for 12 epochs on a 16GB Tesla V100, with a batch size of 8, and a learning rate of 5e-7, via the Transformers library~\citep{wolf2019huggingface}. We use adapted versions of the public\footnote{\url{https://huggingface.co/docs/Transformers/tasks/sequence_classification}} training scripts. Tokenizers pad up to a max length of 256, and the best model by F1 is selected after training. Training took around a day of compute on an NVIDIA A100. We train 25 models stemming from 5 encoders: BERT-base-uncased~\citep{devlin2018bert}, BERT-base-cased, SciBERT cased and uncased~\citep{beltagy2019scibert}, and MathBERT~\citep{shen2021mathbert}. SciBERT is a version of BERT pretrained on scientific text. MathBERT is initialised on BERT-base-uncased, and pretrained on three masked language modelling tasks related to the structure of equation operator trees~\citep{mansouri2019tangent}, and the relationship between equations and their natural language context. It delivers state-of-the-art results in formula search~\citep{zhong2022evaluating}.

\paragraph{Prompting GPT.} For each task, we engineer few-shot prompts with the aim to optimise static performance with respect to the \texttt{gpt-3.5-turbo} model using the OpenAI API. The results of prompt exploration are given in Tab.~\ref{table:prompt_exploration}, where the selected design is highlighted in bold. We describe this prompt below: \newline

\noindent ``The following examples consist of a prompt (denoted by Prompt:) and a label (denoted by Label:).

\texttt{Prompt:} Sequence 1

\texttt{Label:} Label 1

\texttt{Prompt:} Sequence 2

\texttt{Label:} Label 2

\texttt{Prompt:} Sequence 3

\texttt{Label:} Label 3

\texttt{Prompt:} Sequence 4

\texttt{Label:} Label 4\newline
\noindent Now given the following prompt, predict the label.

\texttt{Prompt:} Test Prompt'' \newline

\noindent The sequences all contain the same final annotation as the Test Prompt and are sampled from the training set. Additionally, an equal number of negative (Label: 0) and positive examples (Label: 1) are included as in-context examples, and these examples are shuffled. New lines are denoted by ``\textbackslash n''. Perturbations are only applied to the Test Prompt and in-context examples are fixed to minimise examples' effect on generalisation. 200 random examples from the static test set \textit{per subtask} (e.g. steps=2, integration) are used in the evaluation, which maps to 200 equivalent examples \textit{per perturbation}. This totals around 4000 total examples \textit{per GPT model}.   

{\renewcommand{\arraystretch}{1}%
\begin{table*}[h]
\centering
\scalebox{0.94}{
\begin{tabular}{ c | c | c } 
 \textbf{Prompt Design} & \textbf{GPT-3.5} (F1) & \textbf{GPT-4} (F1)\\ [0.1ex] 
 \hline
 \textit{Derivation Step Classification (steps=2)} & &\\
 No task description + random examples (2 pos, 2 neg) & 61 & 83\\
 Concise task description + random examples (2 pos, 2 neg) & 50 & 83\\
 \textbf{No task description + same final operation examples (2 pos, 2 neg)} & \textbf{68} & \textbf{90}\\
 No task description + same final operation examples (3 pos, 3 neg) & 68 & 87\\
 \hline
 \textit{Calculus Classification} (differentiation) & &\\
 \textbf{No task description (2 pos, 2 neg)} & \textbf{55} & \textbf{55}\\
 No task description (3 pos, 3 neg) & 48 & 64\\
 \hline
\end{tabular}}
\caption{Prompt designs trialled for the experiments.}
\label{table:prompt_exploration}
\end{table*}}

\section{Task-specific data format and sizes}
\label{sec:data}

\begin{figure*}[htp!]
    \centering
    \includegraphics[width=0.95\linewidth]{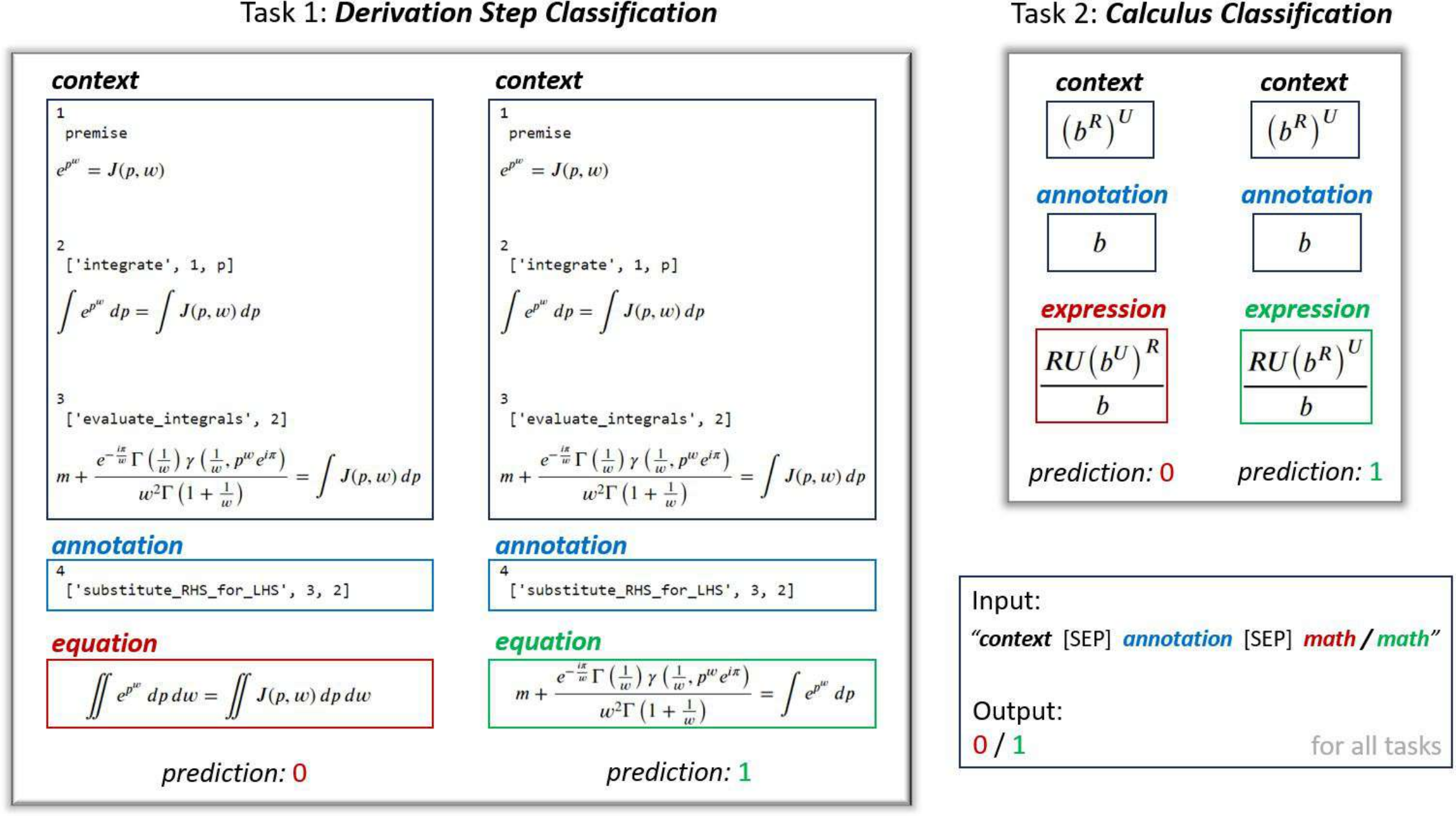}
    \caption{Sampled data from each binary sequence classification task. In short, a sequence containing reasoning \textbf{\textit{context}}, an instruction \textbf{\textit{annotation}}, and resulting \textbf{\textit{math}} is input to a model. The model then predicts whether the math follows from the context and annotation, and if the sequence is mathematically coherent (1) or not (0).}
    \label{fig:input}
\end{figure*}

{\renewcommand{\arraystretch}{1}%
\begin{table*}[h!]
\centering
\scalebox{1}{
\begin{tabular}{ c | c | c | c | c} 
 \textbf{Task} & \textbf{Training} & \textbf{Validation} & \textbf{Static Test} & \textbf{Perturbed Test}\\ [0.1ex] 
 \hline
 \textit{Derivation Step Classification} & & &\\
 2-steps & 20K & 5K & 4K & 4K\\
 3-steps & 20K & 5K & 4K & 4K\\
 4-steps & 20K & 5K & 4K & 4K\\
 \hline
 \textit{Calculus Classification} & & &\\
 integration & 32K & 8K & 4K & 4K\\
 differentiation & 32K & 8K & 4K & 4K\\
 \hline
\end{tabular}}
\caption{The number of examples considered by models during training, validation, and evaluation.}
\label{table:data}
\end{table*}}

The data generation algorithms output a derivation (Alg.~\ref{alg:derivation_generation}) or expression (Alg.~\ref{alg:premise_generation}) in LaTeX and SymPy. Outputs are then adapted to fit specific tasks. For the described classification tasks, a single example consists of the reasoning sequence up to the final expression or equation (Fig.~\ref{fig:input}).

\textbf{Constructing sequences.} For the \textit{Derivation Step Classification} task, a step consists of an equation and an annotation, as described in Fig.~\ref{fig:framework} and Fig.~\ref{fig:input}. An annotation is a list comprising an operator name and its operands. Each step [an, eq] is linearised and comma separated, up to the final step. The final step annotation is separated from the derivation, and the final equation is replaced with a negative example equation, or left unchanged. \newline For \textit{Calculus Classification}, an input sequence consists of a premise expression, a variable, and the result of either differentiating or integrating. The premise expression containing \textit{at least two} variables is initially generated, a variable is randomly selected from the premise, and the resulting expression after differentiating or integrating with respect to that variable is the ground truth. This positive example is either replaced with a negative example, or not. The three main components for each task are [SEP] separated. In the datasets for either task, this sequence is grouped with both the actual final equation and a number of negative equations. As a model encounters an example it is processed into two sequences; one including the positive equation and another including the negative. Each sequence is then paired with the corresponding classification labels. Perturbations are applied to each test set and generate an equal number of perturbed examples. The \textit{Derivation Step Classification} datasets include 41K evaluation examples \textit{per derivation step count}. The \textit{Calculus Classification} datasets include 52K evaluation examples \textit{per operation}. This equates to \textbf{\textit{227K total examples}}. Tab.~\ref{table:data} describes the relevant sizes for the models.

\section{Supplementary Material for Qualitative Analysis}
\label{sec:qual_mat}

We can alternatively measure generalisability by examining the proportion of examples where predictions involving static sequences are correct, while predictions for mathematically equivalent perturbed sequences are incorrect. Defining an example to consist of a static sequence grouped with its perturbed equivalents, if a static prediction is correct while all perturbation predictions fail, this gives a strict measure of generalisability  (denoted by $G$ in Tab.~\ref{table:pairwise_analysis}) and complements previous analysis. These grouped examples allow examination of how well models understand each operator, and can highlight their weaknesses. We identify such weaknesses shared between GPT and BERT models and discuss clear dissimilarities in a more focused discussion in this section. 

\textbf{Why is $R$ associated with generalisation failure for BERT but not for GPT?} Prior analysis points to the premise renaming operator $R$ as a useful point of comparison between fine-tuned BERT and few-shot GPT. Prompting GPT-3.5 by appending ``Describe what function \texttt{renaming\_premise} performs.'' to a static prompt (associated with GPT-3.5's generalisation failure) returns the following definition of $R$: \textit{``the} \texttt{renaming\_premise} \textit{function is used to create a new expression or equation by assigning an existing expression or function to a new variable or function symbol.''} This appropriate understanding persists even for perturbed prompts, and naturally extends to GPT-4. In contrast, further analysis (Appendix~\ref{sec:qual_mat}) reinforces that BERT models do not share this out-of-distribution understanding. The main difference between $R$ and all other operators is that it appears in sequences \textit{without any reference to prior equations}. The substitution operations are the opposite of this (referencing the most equations of any operators), and \textit{both} GPT-4 and BERT frequently fail to make correct predictions given this operator. On one hand, the operator with the \textit{least referencing} is significantly associated with generalisation failure for BERT, but not GPT-4. On the other, the operator with the \textit{most referencing} is not significantly associated with generalisation failure in either case, as all models are not effectively learning substitution in-distribution. BERT is dependent on more localised learning where the necessary semantics is expressed within a short text span during training, rather than a span that explicitly relates to other textual elements (\textit{e.g.,} through regular reference). In other words, \textit{a lack of explicit discourse relations that predictably vary with the ground truth obstructs models from learning latent relations that allow them to generalise.} However, the explicit relations can not be too complex (as with substitution). $R$ lends itself to generalisation failure because it lacks structured discourse relations of the appropriate complexity for BERT (that others operators do not). $R$ is simpler for GPT because of its varied exposure to structured text featuring such relations (\textit{e.g.,} code) and obviously its relative size. \newline

\noindent Focusing solely on BERT-related models in more depth, we consider (uncased) models trained on 3-step derivations. This number of steps closely reflects the average results over all step counts in Table~\ref{tab:steps}. The \textbf{All} (perfect generalisation) and \textbf{Not P} (complete generalisation failure) columns of Table~\ref{table:qual_1} (Appendix~\ref{sec:qual_mat}) reinforce the relative generalisability gap between SciBERT and MathBERT, despite both being trained on scientific corpora, and display the top three operators by normalised frequency per generalisation category. 

\begin{figure}[t]
    \centering
    \includegraphics[width=1\columnwidth]{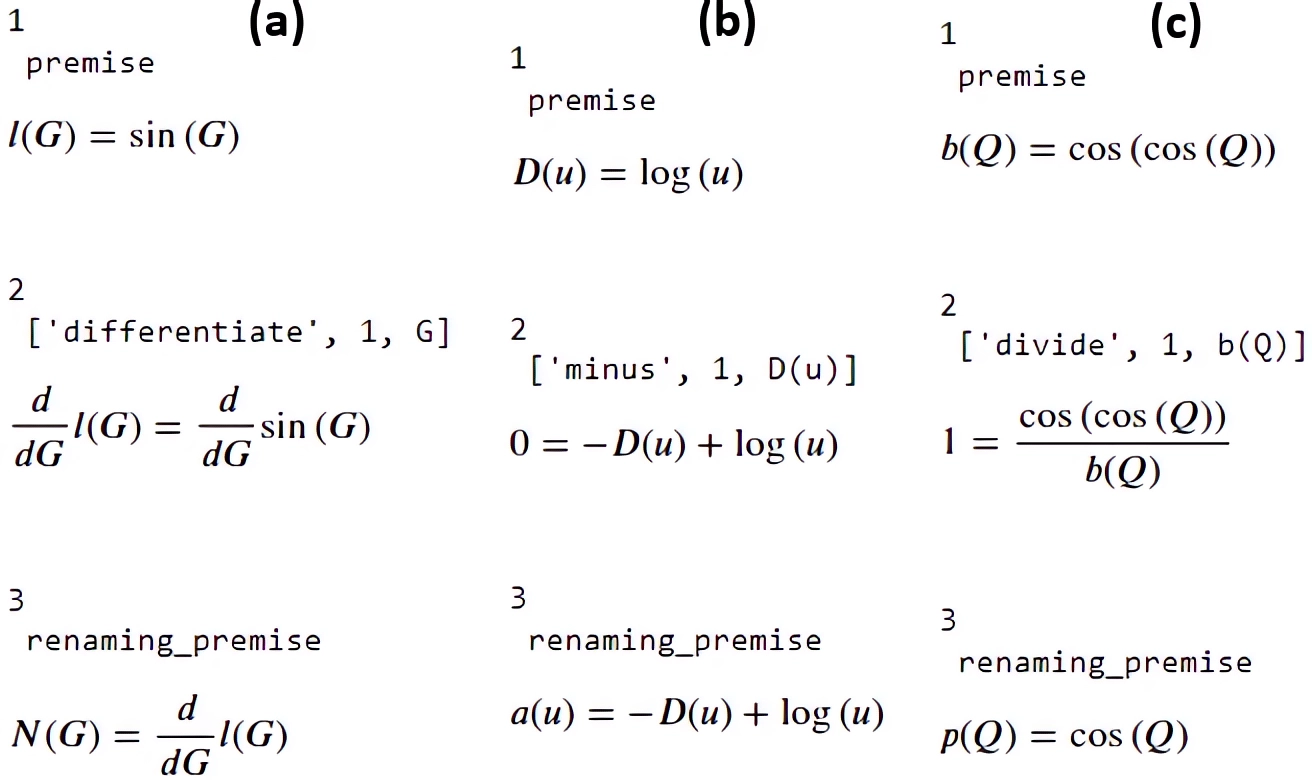}
    \caption{Three examples of the total 15 where both SciBERT and MathBERT correctly classify \textit{unperturbed} examples (as shown), but incorrectly classify all perturbed examples.}
    \label{fig:bad_examples}
\end{figure}

\textbf{Generalisation failure depends on the unpredictability of an operator.} For examples where models perfectly generalise, the operator responsible for setting up an integral (without evaluating it) is most common. This is likely because it involves prepending a unique text span "\textbackslash int" to expressions either side of equations, which is easy to identify. Models generalise well to $\cos$, $\sin$, $\exp$, and $\log$ operators, likely due to their similarly predictable effect on equations associated with regular text spans. To highlight that it is likely the relative unpredictability of an operator's effect on text that leads to generalisation failure, we analyse the set of examples where \textit{both} SciBERT and MathBERT correctly classify unperturbed sequences, but misclassify \textit{all} perturbed sequences. Three examples are displayed in Fig.~\ref{fig:bad_examples}. The \textit{renaming premise} operation is overwhelmingly frequent. It takes a random previously defined expression as the RHS, and defines a new function as the LHS. It does not necessarily depend on a single previous step and is non-deterministic due to random sampling of the RHS, yet it can never generate more complex equations than those previously derived (unlike other operators).

\textbf{Entailment pre-training improves generalisability.} BERT~\citep{devlin2018bert} was trained on masked language modelling (MLM) and next sentence prediction (NSP) objectives. SciBERT~\citep{beltagy2019scibert} was further trained with scientific papers on MLM and NSP. MathBERT~\citep{shen2021mathbert} was further trained from BERT on educational mathematical text, ranging from pre-k to graduate level difficulty. However, unlike BERT and SciBERT, MathBERT was trained to optimise performance on MLM over a large corpus. Fine-tuning generally overwrites representations learned from previous tasks~\citep{mosbach2020stability}, and MathBERT has likely forgotten those associated with NSP. The current classification tasks involve determining if \textit{math context entails an expression/equation}, rather than predicting individual tokens as in language modelling. Next-equation prediction shares greater similarity with NSP than MLM, and we therefore attribute generalisability failures of MathBERT in this context to insufficient entailment pre-training. It has struggled with entailment before relative to other BERT models~\citep{meadows2022physnlu}.

\textbf{Advantages of pre-training on structured scientific text.} SciBERT differs from the other encoders due to a distinct focus on scientific papers written in LaTeX. This offers two benefits: \textbf{(1)} Mathematical elements seen by models are written in LaTeX, so exposure to LaTeX (during both MLM \textit{and} NSP) provides natural advantage; \textbf{(2)} Scientific papers tend to be concise and logically structured. Text spans are carefully chained to reach conclusions, so exposure to papers during training may better teach models the concept of entailment and aid performance in related tasks. 

{\renewcommand{\arraystretch}{1.3}%
\begin{table}[h]
\centering
\small
\scalebox{1}{
\begin{tabular}{ c | c | c | c } 
  & \textbf{Static $\pm$} & \textbf{All} & \textbf{Not P}\\ [0.1ex] 
 \hline
 \multirow{2}{*}{BERT} & 62.3 & 7.4 & 5.3\\
 & $R \; \int_E \; \partial_E$ & $\partial_E \; \int \; -$ & $S_L \; \int_E \; R$\\
 \hline
 \multirow{2}{*}{SciBERT} & 79.6 & 21.3 & 1.6 \\
 & $R \; \int_E \; \partial_E$ & $\int \; \partial_E \; \cos$ & $R \; X^O \; \times$\\
 \hline
 \multirow{2}{*}{MathBERT} & 70.3 & 7.8 & 9.3\\
 & $R \; \int_E \; \int $ & $\int \; \cos \; \sin$ & $R \; \partial_E \; \int_E$\\
 \hline
\end{tabular}}
\caption{
\textbf{Static $\pm$} is the rate at which positive and associated negative \textit{unperturbed} sequences are \textit{both} correctly classified. \textbf{All} (perfect generalisation) is the percentage of examples where the static and perturbed (positive and negative) sequences are correctly classified. \textbf{Not P} (complete failure to generalise) is percentage of examples where only the static positive sequences are classified correctly, while all perturbed positive sequences are incorrect. Symbols correspond to the top three most frequent (final) operators in each unperturbed sequence, where frequency is normalized with respect to operator frequency in the static set. $R$ is a premise renaming operator. $\int$ and $\partial$ are integration and differentation operators. $\int_E$ and $\partial_E$ are respective evaluation operators. $X^O$ is exponentiation, $\times$ is multiplication, $-$ is subtraction, and $S_L$ is LHS substitution.}
\label{table:qual_1}
\end{table}}

\section{Algorithm for Premise Generation}
\label{sec:premise_generation}

The \textit{"Generate Premise Equation"} algorithm (Alg.\ref{alg:premise_generation}) aims to create a mathematical equation from a defined vocabulary of letters and operators. Specifically, the algorithm's process can be summarized as follows:

\begin{enumerate}
    \item \textbf{Initialisation:} Symbols and mathematical operations are defined:
    \begin{itemize}
        \item \(\mathcal{S}\) represents all symbols from the vocabulary \(\mathcal{V}\).
        \item \(\mathcal{R}_1\) comprises unary operations like Cosine, Sine, Exponential, and Logarithm.
        \item \(\mathcal{R}_2\) contains binary operations such as Addition, Subtraction, Multiplication, etc.
    \end{itemize}

    \item \textbf{Base RHS Construction:} Depending on a randomly chosen arity (either 1 for unary or 2 for binary):
    \begin{itemize}
        \item If arity = 1, the RHS is built by applying a random unary operation on a random symbol.
        \item If arity = 2, the RHS is constructed using a binary operation on two distinct random symbols.
    \end{itemize}

    \item \textbf{Complexifying the RHS:} A random complexity value is selected from 0 to \(\mathcal{C}-1\). For each iteration up to the chosen complexity, the RHS's complexity is increased by applying either a unary operation on the current RHS or a binary operation between the current RHS and another random symbol.

    \item \textbf{LHS Construction:} The LHS is then formulated as a function of the free symbols present in the RHS.

    \item \textbf{Equation Formation:} Lastly, an equation, termed \textit{premise}, is crafted using the finalized LHS and RHS.
\end{enumerate}

In essence, this algorithm dynamically produces a mathematical equation whose intricacy varies depending on the randomly chosen operations and the selected complexity.

\begin{algorithm*}
Assumes a global vocabulary of letters, $\mathcal{V}$ and operators \textit{e.g.,} cos, sin, etc. Accepts a complexity hyperparameter $\mathcal{C}$ that determines the maximum tree depth of the premise RHS.
\caption{\textbf{Generate Premise Equation}}
\begin{algorithmic}[1]
\Procedure{Premise}{$\mathcal{C}$}
\State $\mathcal{S} \gets$ [Symbol($v$) for $v$ in $\mathcal{V}$]
\State $\mathcal{R}_1 \gets$ [Cos, Sin, Exp, Log]
\State $\mathcal{R}_2 \gets$ [Add, Minus, Times, Power, Divide, Differentiate, Integrate]
\State arity $\gets$ random.choice([1,2])
\If {arity $=$ 1}
\State $R \gets$ random.choice($\mathcal{R}_1$)
\State $S \gets$ random.choice($\mathcal{S}$)
\State RHS $\gets$ $R(S)$
\State LHS $\gets$ random.choice([$s$ for $s$ in $\mathcal{S}$ if $s$ $\neq S$])
\ElsIf {arity $=$ 2}
\State $R \gets$ random.choice([$r$ for $r$ in $\mathcal{R}_2$ if $r$ not in [Differentiate, Integrate]])
\State $S_1 \gets$ random.choice($\mathcal{S}$)
\State $S_2 \gets$ random.choice([$s$ for $s$ in $\mathcal{S}$ if $s \neq S_1$])
\State RHS $\gets R(S_1, S_2)$
\State LHS $\gets$ random.choice([$s$ for $s$ in $\mathcal{S}$ if $s$ not in [$S_1$, $S_2$]])
\EndIf
\State complexity $\gets$ random.choice(range($\mathcal{C}$))
\For {$i \in$ range(complexity)}
\State arity $\gets$ random.choice([1,2])
\If {arity $=$ 1}
\State $R \gets$ random.choice($\mathcal{R}_1$)
\State RHS $\gets R($RHS$)$
\ElsIf {arity $=$ 2}
\State $R \gets$ random.choice($\mathcal{R}_2$)
\State $S \gets$ random.choice($\mathcal{S}$)
\State RHS $\gets R($RHS$, S)$
\EndIf
\EndFor
\State LHS $\gets$ Function(LHS)(*tuple(RHS.free\_symbols))
\State premise $\gets$ Eq$($LHS, RHS$)$
\State \textbf{return} premise
\EndProcedure
\end{algorithmic}
\label{alg:premise_generation}
\end{algorithm*}

\section{Algorithm for Derivation Generation}
\label{sec:derivation_generation}

Algorithm~\ref{alg:derivation_generation} relies on Algorithm~\ref{alg:premise_generation} in order to derive subsequent equations. It relies on two other procedures other than Step. The EquationDistribution function relies on the hyperparameter $p_h$, which controls the frequency that recent equations are sampled as a cubic function of $p_h$. The ExtractDerivation function is responsible for collecting all related steps from the initial longer derivation, such that a final self-contained derivation is obtained. This derivation must match the desired length, $L_f$. 

\paragraph{Runtime.} To calculate the time taken to generate a derivation, we sample a number of derivation lengths (\textit{i.e.,} number of equations) from a Gaussian $N(6.5, 3)$ truncated between $4$ and $9$ inclusive. It took $71$ minutes to generate $100$ derivations, with an average length of $6$ (\textit{i.e.,} $0.7$ min/derivation, $7$ seconds per step), on a mid-range laptop CPU. In addition to the time taken for SymPy to perform complex calculus operations, it takes some time to generate valid derivations that do not repeat steps and fit within our given parameters, hence hundreds of steps may fail for a given derivation.

\paragraph{Hyperparameters.} We rely on other hyperparameters to control \textbf{1.} the selection bias towards operations being applied to more recent equations, \textbf{2.} the bias towards operators of a particular arity, and \textbf{3.} bias towards other operator subcategories. \newline Considering \textbf{1.}, in the 2-arity two annotation format [`operator', operand 1, operand 2], operand 1 is always an equation index. This is also true for 1-arity, and 0-arity does not require an operand. An equation is randomly sampled from a non-repeating set of derived equations. The \textit{history hyperparameter}, $p_h$, clones an equation in the list through a cubic function of its step-wise chronological position as described above. With our default settings, the last equation in a list of three is twice as likely to be selected as input than the first. This emulates mathematicians working with recent equations, but having to occasionally sample from distant results. \newline Other hyperparameters work similarly, by repeating elements of lists. Considering \textbf{2.}, we bias towards 2-arity, as those contain calculus, and considering \textbf{3.} we bias towards substitution operations, differentiation, and integration. The exact form of the algorithm used to generate data for this paper is available in the linked repository on the first page. 

In more formal detail, the mechanics of Algorithm~\ref{alg:derivation_generation} are as follows:

\begin{enumerate}
    \item \textbf{Procedure Step}: This subroutine generates a single step in the derivation.
    \begin{itemize}
        \item Sets of equations, operations, and other relevant elements are initialised from the dataset \(\mathcal{D}\).
        \item Based on probability parameters, the arity of the operation (either 0, 1, or 2) for this step is determined.
        \item Depending on the chosen arity:
        \begin{itemize}
            \item Arity 0: The equation and annotation for this step are directly chosen from the set \(\mathcal{R}_0\).
            \item Arity 1: An operation from \(\mathcal{R}_1\) and an equation from the dataset are chosen to form the new equation.
            \item Arity 2: An operation from \(\mathcal{R}_2\), an equation from the dataset, and another element are selected to shape the equation.
        \end{itemize}
        \item If the formed equation is deemed valid through certain checks it is returned; otherwise, None is returned.
    \end{itemize}

    \item \textbf{Main Derivation Loop}: This section assembles the derivation.
    \begin{itemize}
        \item The initial step of the derivation is generated using Algorithm~\ref{alg:premise_generation}.
        \item A pre-defined target length \(L_i\) describes approximately the number of times the \textit{Step} procedure is invoked to add new steps.
        \item The full derivation is extracted from the accumulated steps.
        \item The loop breaks when the derivation reaches a desired length \(L_f\), where \(L_f \geq L_i\).
    \end{itemize}
\end{enumerate}

To summarize, the algorithm iteratively constructs a derivation of mathematical equations, where each step is shaped by a series of operations determined by specific probabilities and conditions. It is given on the following page.

\begin{algorithm*}
\caption{\textbf{Generate Equational Reasoning}}
\begin{algorithmic}[1]

\Procedure {Step}{$\mathcal{D}, p_0, p_1, p_2, p_h, p_r, p_e, p_c, p_s$}
\State $D \gets$ [$i[0]$ for $i$ in $\mathcal{D}$]
\State $A \gets$ [$i[1]$ for $i$ in $\mathcal{D}$]
\State $\mathcal{R}_0 \gets$ [Premise] + [RenamingPremise]$\times p_r$
\State $\mathcal{R}_1 \gets$ [Cos, Sin, Exp, Log, Expand] + [EvaluateDerivatives, EvaluateIntegrals]$\times p_e$
\State $\mathcal{R}_2 \gets$ [Add, Minus, Times, Divide, Power] + [Differentiate, Integrate]$\times p_c$ \newline \indent \indent + [SubsLHSForRHS, SubsRHSForLHS]$\times p_s$
\State elements $\gets$ numbers, variables, and subexpressions from $D$
\State arity $\gets$ random.choice([0]$\times p_0$ + [1]$\times p_1$ + [2]$\times p_2$)
\If {arity $=$ 0}
\State $R \gets$ random.choice($\mathcal{R}_0$)
\State equation $\gets R$
\State annotation $\gets R$.\_\_name\_\_
\ElsIf {arity $=$ 1}
\State $R \gets$ random.choice($\mathcal{R}_1$)
\State $e_1 \gets$ random.choice(EquationDistribution($D, p_h$))
\State equation $\gets R(e_1)$
\State $n \gets D$.index($e_1$)
\State annotation $\gets$ [$R$.\_\_name\_\_, $n+1$]
\ElsIf {arity $=$ 2}
\State $R \gets$ random.choice($\mathcal{R}_2$) \Comment{$R$ depends on the length of $D$}
\State $e_1 \gets$ random.choice(EquationDistribution($D, p_h$))
\State $e_2 \gets$ random.choice(elements) \Comment{$e_2$ will vary depending on $R$}
\State equation $\gets R(e_1, e_2)$
\State $n \gets D$.index($e_1$)
\State annotation $\gets$ [$R$.\_\_name\_\_, $n+1$, $e_2$]
\EndIf
\If {equation is valid} \Comment{validity depends on various checks}
\State \textbf{return} equation, annotation
\Else
\State \textbf{return} None
\EndIf
\EndProcedure
\While {True}
\State $\mathcal{D} \gets$ [(Premise($\mathcal{C}$), "premise")] \Comment{generate first step using Algorithm~\ref{alg:premise_generation}}
\While {len($\mathcal{D}$) $< L_i$} \Comment{$L_i$ is an initial length of the derivation}
\State step $\gets$ Step($\mathcal{D}, p_0, p_1, p_2, p_h, p_r, p_e, p_c, p_s$)
\If {step is not None}
\State $\mathcal{D}$.append(step)
\EndIf
\EndWhile
\State derivation $\gets$ ExtractDerivation($\mathcal{D}$)
\If {len(derivation) $= L_f$} \Comment{$L_f \geq L_i$ is the desired length of the derivation}
\State \textbf{break}
\EndIf
\EndWhile
\State $\mathcal{D} \gets$ derivation

\end{algorithmic}
\label{alg:derivation_generation}
\end{algorithm*}



\end{document}